\documentclass[11pt]{article}

\usepackage[preprint]{acl}

\usepackage{times}
\usepackage{latexsym}
\usepackage[T1]{fontenc}
\usepackage[utf8]{inputenc}
\usepackage{microtype}
\usepackage{inconsolata}
\usepackage{graphicx}
\usepackage{float}
\usepackage{amsmath}
\usepackage{amssymb}
\usepackage{booktabs}
\usepackage{multirow}
\usepackage{xcolor}
\usepackage{subcaption}
\usepackage{algorithm}
\usepackage{algorithmic}

\newcommand{\styleshield}{\textsc{DIAL}}

\newcommand{\pai}{P_{\mathrm{AI}}}

\title{Does It Appear Less AI-Generated? Using Flow Matching to Test Detectors Across a Continuum of Rewrite Strengths}

\author{Guantian Zheng\thanks{Corresponding author: \href{mailto:gzheng004@e.ntu.edu.sg}{\textnormal{gzheng004@e.ntu.edu.sg}}} \\
  {\fontfamily{lmtt}\fontseries{l}\fontsize{10}{12}\selectfont Automan Lab, Nanyang Technological University}}

\hypersetup{pdfauthor={Guantian Zheng}}

\begin{document}
\setcounter{footnote}{1}
\maketitle

\begin{abstract}
When a commercial AI detector labels a paper ``30\% AI-generated,'' what does that number tell us about the quality of the work or the care that went into writing it?
As language models learn to reproduce human prose, style may become a less reliable guide to authorship, yet detector scores can carry weight in academic integrity reviews.
We introduce DIAL, a flow-matching model for testing how detector judgments change with rewriting strength.
Conditioned on frozen Qwen2.5-7B representations, DIAL rewrites a passage in token embedding space. A single parameter $\gamma$ sets the initial noise level, giving continuous control over rewriting strength.
On a multi-domain Chinese benchmark, evasion against the training detector rose from 52.6\% at $\gamma{=}6.0$ through 79.7\% at $\gamma{=}6.5$ to 94.6\% at $\gamma{=}7.0$. At the latter setting, evasion reached at least 99\% against each of three unseen detectors, with a Qwen embedding similarity of 0.928 to the source.
Readers rated these rewrites as more human-like but less faithful on average than those at $\gamma{=}6.5$.
RateAudit selectively rewrites passages to steer document scores toward six targets from 10\% to 60\%. A 30\% target yielded a mean score of 30.5\% across 50 long documents, illustrating how stylistic changes can reshape detection percentages used in academic evaluation.
We release our code and data at \url{https://github.com/Ethan-Zheng136/DIAL}.
\end{abstract}

\section{Introduction}
\label{sec:intro}

Should a student be penalized when a commercial detector flags a large share of their essay as AI-generated?
Detectors infer the origin of a text from the finished prose~\citep{he2024mgtbench,yang2024survey}. An instructor assessing that essay must also judge the quality of its argument and the student's understanding of the material.
Before a detection report can carry weight in that judgment, we need to understand how much its verdict depends on the way the essay is written.

The current detector paradigm stands on shaky ground.
The stylistic boundary between AI-generated and human-written text is increasingly difficult to draw.
LLMs learn from human-written corpora, and writers increasingly use these models to revise their own prose.
The stylistic features that detectors rely on can therefore appear in both kinds of text.
Commercial incentives compound the problem.
In China, some universities use commercial detectors to screen theses, and some of the same providers also sell rewriting services to lower the scores they report~\citep{restofworld2025aigc}.
A flagged passage can thus create demand for another service from the very provider that flagged it.
This is one reason we conduct our evaluation in Chinese.
\citet{sadasivan2023aigenerated} show that even when the source of a text is known, repeated paraphrasing can substantially reduce detector performance.
Taken together, these observations make it difficult to justify using a detector's estimate of a text's origin to judge the quality of the work.

To examine this sensitivity, we evaluate three common approaches---LLM-based style transfer, backtranslation, and synonym substitution---on Chinese text.
LLM-based style transfer can retain patterns that detectors associate with AI generation, yielding little improvement in evasion.
Backtranslation reduces detector scores at the cost of semantic fidelity, while synonym substitution makes surface-level changes that barely affect detector confidence.
These outcomes motivate continuous control over style-transfer strength to examine the trade-off between changing a detector's judgment and preserving the original content.
By progressively varying the style of the same source passage, we can trace how far the detector score shifts.

We propose DIAL, a flow-matching framework for continuously controllable text style transfer in token embedding space.
Adapting the SDEdit paradigm~\citep{meng2022sdedit}, we add noise to the source embeddings and denoise them with a pretrained generative model.
A single inference parameter, $\gamma$, sets the initial noise level, providing continuous control over transfer strength.
We pretrain a DiT backbone~\citep{peebles2023scalable,chen2026langflow} with flow matching on Chinese text.
To condition denoising on the source passage, we use cross-attention adapters with zero-initialized gates to read hidden states from a frozen Qwen2.5-7B-Instruct model.
During training, detector scores provide an auxiliary reward.

\noindent Contributions.\quad
\begingroup
\setlength{\topsep}{4pt}
\setlength{\partopsep}{0pt}
\begin{enumerate}
\setlength{\itemsep}{2pt}
\setlength{\parsep}{0pt}
    \item We introduce DIAL, a source-conditioned flow-matching model for continuously controllable text style transfer, serving as an adversarial probe that exposes systematic vulnerabilities across multiple AIGC detectors.

    \item We construct a corpus of 436,000 paired Chinese texts spanning social media, news, and academic abstracts.
    On 1,000 held-out examples, we trace how detector scores and semantic similarity vary with transfer strength and evaluate generalization to three detectors held out from training.

    \item We introduce RateAudit, a document-level diagnostic that selectively applies style transfer to address a central question.
    If a detector reports $x\%$ AI-generated content, can that figure be steered toward a chosen target?
    Our answer is yes, calling into question the reliability of detection-rate-based verdicts.
\end{enumerate}
\endgroup

\section{Related Work}
\label{sec:related}

\paragraph{AIGC Detection.}
Supervised AIGC detectors, typically BERT-based classifiers trained
on human/machine text corpora~\citep{devlin2019bert, he2024mgtbench},
are complemented by zero-shot methods exploiting probability
curvature~\citep{mitchell2023detectgpt} or
watermarking~\citep{kirchenbauer2023watermark}.
However, \citet{sadasivan2023aigenerated} show that simple
paraphrasing dramatically reduces detection accuracy, and
cross-domain transfer remains
fragile~\citep{he2024mgtbench, yang2024survey}.

\paragraph{Text Style Transfer and Adversarial Probing.}
Adversarial approaches to text classification range from lexical substitution to learned text transformations.
TextFooler~\citep{jin2020bert} changes classifier predictions through token substitutions.
DIPPER~\citep{krishna2024paraphrasing} evades AIGC detectors through paragraph-level paraphrasing, with discrete control levels for lexical diversity and content reordering.
Adversarial Paraphrasing~\citep{cheng2025adversarial} uses detector feedback to guide an instruction-following LLM, while MASH~\citep{gu2026mash} combines supervised fine-tuning, preference optimization, and inference-time refinement for style humanization.
These approaches do not establish a continuous diagnostic axis over transformation intensity---a capability naturally supported by continuous embedding-space generation.

\paragraph{Diffusion and Flow Matching for Text.}
\citet{li2022diffusion} propose Diffusion-LM for controllable generation in embedding space.
\citet{lou2024discrete} develop SEDD, a discrete diffusion model for text.
\citet{chen2026langflow} show that continuous flow matching rivals discrete diffusion in language modeling.
ParaGuide~\citep{horvitz2024paraguide} performs paraphrase-conditioned diffusion for style transfer, with a guidance coefficient controlling the trade-off between style and semantic preservation.
DIAL combines source-conditioned flow matching with SDEdit-style inference, controlling transfer strength through the initial noise level applied to source embeddings.

\section{Method}
\label{sec:method}

\begin{figure}[!t]
    \centering
    \includegraphics[width=\columnwidth]{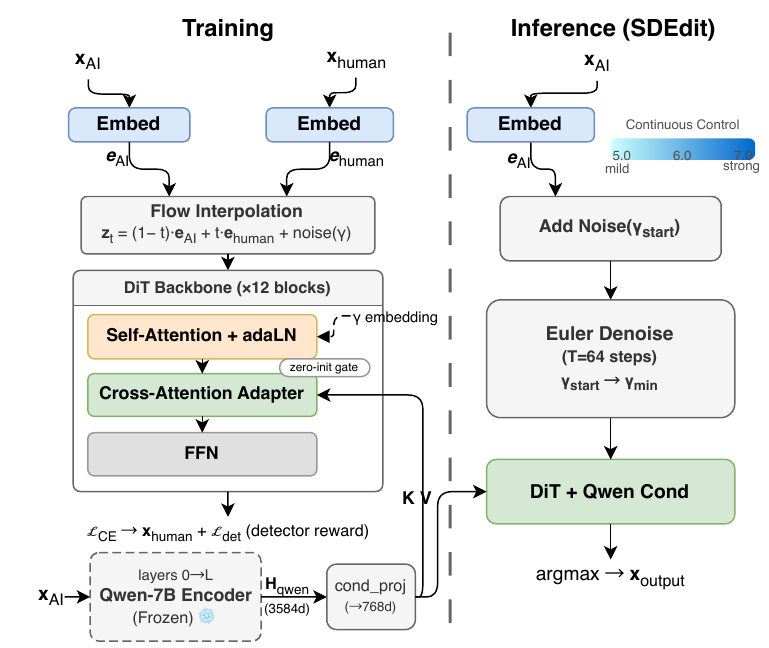}
    \caption{Overview of \styleshield{}.
    A pretrained DiT backbone (left) is augmented with cross-attention
    adapters with zero-initialized gates. These adapters attend to frozen
    Qwen2.5-7B hidden representations (bottom).
    At inference (right), AI text embeddings are noised at level~$\gamma$
    and iteratively denoised with semantic conditioning via 64 Euler steps.
    The single parameter~$\gamma$ serves as a continuous
    diagnostic axis for characterizing the
    evasion--preservation trade-off.}
    \label{fig:architecture}
\end{figure}

We present \styleshield{}, a conditional flow-matching framework
that operates in continuous token-embedding space, designed as
an adversarial diagnostic for probing AIGC detector robustness.
The system has three components: a DiT backbone pretrained with
flow matching (\S\ref{sec:backbone}), cross-attention adapters
conditioned on a frozen Qwen encoder
(\S\ref{sec:conditioning}), and a training pipeline with
detector-in-the-loop feedback (\S\ref{sec:training}).
At inference a single parameter~$\gamma$ provides continuous
control over transformation intensity, serving as a diagnostic
axis for characterizing detector sensitivity
(\S\ref{sec:inference}).
We further describe RateAudit, a document-level
scheduling algorithm that stress-tests whether detection-rate
verdicts can be set to arbitrary values
(\S\ref{sec:allocator}).
Figure~\ref{fig:architecture} shows the overall architecture.

\suppressfloats[t]

%% ────────────────────────────────────────────────────
\subsection{Flow Matching Backbone}
\label{sec:backbone}

DIAL operates in continuous token-embedding space using
a Diffusion Transformer (DiT;~\citealp{peebles2023scalable})
pretrained with flow matching~\citep{lipman2023flow} on Chinese text.
We augment this backbone with source-conditioned cross-attention
adapters (\S\ref{sec:conditioning}) and further train it for
style transfer (\S\ref{sec:training}).
Each token~$w_i$ is mapped to a $d$-dimensional embedding
$\mathbf{e}_i \in \mathbb{R}^d$ ($d{=}768$) via a learned
vocabulary embedding layer.
Embeddings are length-normalized to stabilize training in the
continuous space:
$\bar{\mathbf{e}} = \sqrt{d}\;\mathbf{e}/\|\mathbf{e}\|$.

The DiT consists of $N{=}12$ transformer blocks with
self-attention (RoPE), FFN (GELU, expansion ratio~4), and
adaLN-Zero conditioning.
A timestep embedder maps~$\gamma$ to
$\mathbf{c}_\gamma$, which modulates each block:
\begin{equation}
    \mathbf{x}'
      = (1 + \mathbf{s}_\gamma)
        \odot \mathrm{LN}(\mathbf{x})
        + \mathbf{b}_\gamma,
\end{equation}
where $\mathbf{s}_\gamma, \mathbf{b}_\gamma$ are predicted from
$\mathbf{c}_\gamma$ via a zero-initialized linear layer.
We additionally employ a bias skip
connection~\citep{karras2022elucidating}:
\begin{equation}
    \mathrm{logits}
      = f_\theta(\mathbf{z},\gamma)
        + c_{\mathrm{skip}}(\gamma)
          \,\mathbf{z}\,\mathbf{W}_{\mathrm{emb}}^\top,
\end{equation}
where
$c_{\mathrm{skip}}(\gamma)
  = \exp\bigl((\mathrm{softplus}(-\gamma)-\gamma)/2\bigr)$
provides a direct residual path from the noisy input to the
output logits, stabilizing training at low noise levels.

We adopt a variance-preserving (logit-SNR) noise
schedule: the forward noising process at noise
level~$\gamma$ is:
\begin{equation}
    \mathbf{z}_\gamma
      = \alpha(\gamma)\,\mathbf{x}
      + \sigma(\gamma)\,\boldsymbol{\epsilon},
    \quad
    \boldsymbol{\epsilon}\sim\mathcal{N}(\mathbf{0},\mathbf{I}),
    \label{eq:noise}
\end{equation}
where $\alpha(\gamma) = \sqrt{\mathrm{sigmoid}(-\gamma)}$ and
$\sigma(\gamma) = \sqrt{\mathrm{sigmoid}(\gamma)}$.
During training, noise levels are sampled from a learned Gumbel
proposal
$\gamma \sim \mathrm{Gumbel}(\mu, \beta)$.

%% ────────────────────────────────────────────────────
\subsection{Cross-Attention Conditioning}
\label{sec:conditioning}

Style transfer, unlike unconditional generation, must preserve
the semantic content of a specific input text.
We inject this content signal via cross-attention adapters
conditioned on a frozen large language model.
We use Qwen2.5-7B-Instruct~\citep{qwen2024qwen25} as a frozen
semantic encoder for its Chinese-language capabilities and
accessible intermediate representations.
The AI-text input is tokenized with the Qwen tokenizer and
passed through layers~$0$ to~$L{-}1$
(default $L{=}14$ of 28); the resulting hidden states
$\mathbf{H}_{\mathrm{qwen}}
  \!\in\! \mathbb{R}^{B \times S_q \times 3584}$
encode the semantic content to be preserved.
A learned projection
$\mathrm{proj}\!:\!\mathbb{R}^{3584}{\to}\mathbb{R}^{768}$
maps Qwen features to the DiT hidden dimension.
The Qwen model is kept entirely frozen throughout training---no
gradients flow through it.
The choice of split layer~$L$ trades surface-level fidelity
against semantic abstraction; we ablate
$L\!\in\!\{7,14,21\}$ in \S\ref{sec:ablation}, finding
$L{=}14$ optimal.

After each DiT self-attention block, we insert a
cross-attention adapter.
Each adapter performs multi-head cross-attention, with queries
computed from the DiT hidden states and keys and values from the
projected Qwen hidden states:
\begin{align}
    \mathbf{Q} &= \mathbf{W}_q\,\mathrm{LN}(\mathbf{x}),
    \;
    [\mathbf{K};\mathbf{V}]
      = \mathbf{W}_{kv}\,\mathrm{proj}(\mathbf{H}_{\mathrm{qwen}}),
    \label{eq:cross-kv}\\
    \mathbf{x}'
      &= \mathbf{x}
       + g(\mathbf{c}_\gamma)\odot
         \mathrm{MHA}(\mathbf{Q},\mathbf{K},\mathbf{V}).
    \label{eq:cross-gate}
\end{align}
The gate
$g(\mathbf{c}_\gamma)
  = \mathbf{W}_g\,\mathbf{c}_\gamma + \mathbf{b}_g$
is initialized to zero
($\mathbf{W}_g{=}\mathbf{0},\;\mathbf{b}_g{=}\mathbf{0}$).
This zero initialization ensures the model begins as the exact
pretrained backbone ($g \equiv \mathbf{0}$) and smoothly learns
to incorporate conditioning without catastrophic forgetting.
The cross-attention adapters add approximately 28M parameters in total.

%% ────────────────────────────────────────────────────
\subsection{Training}
\label{sec:training}

We construct 436K parallel (AI, human) text pairs across three
Chinese-language domains---social media (346K), news (68K), and
academic abstracts (22K).
For each domain, we collect human-written texts and prompt
Qwen2.5-7B-Instruct to rewrite them into AI style.
Each pair is then scored by Det-v3 and subjected to
dual-end strict filtering: we retain pairs with
$\pai < 0.1$ on the human side and $\pai > 0.9$ on the AI side
to provide contrasting training endpoints under Det-v3.

To train the model for style transfer, we embed both texts in
each pair $(x_{\mathrm{AI}}, x_{\mathrm{hu}})$, sample
$t\!\sim\!\mathcal{U}(0,1)$ and~$\gamma$
from the Gumbel proposal, form the noisy input via flow
interpolation
$\mathbf{z}
  = (1{-}t)\,\mathbf{e}_{\mathrm{AI}}
  + t\,\mathbf{e}_{\mathrm{hu}}$
followed by noise corruption (Eq.~\ref{eq:noise}), and
minimize the denoising cross-entropy:
\begin{equation}
    \mathcal{L}_{\mathrm{CE}}
      = -\!\sum_{i=1}^{S}
          \log\, p_\theta\!\bigl(
            w_i^{\mathrm{hu}}
            \;\big|\;
            \mathbf{z}_\gamma,\; \gamma,\;
            \mathbf{H}_{\mathrm{qwen}}
          \bigr).
    \label{eq:ce}
\end{equation}
The Qwen condition $\mathbf{H}_{\mathrm{qwen}}$ is computed
from~$x_{\mathrm{AI}}$; thus the model learns to denoise
toward the human target while attending to the source semantics.

After a 5{,}000-step warmup we introduce an auxiliary
adversarial signal from a frozen BERT-based AIGC detector.
The full objective adds a detector term to the denoising
loss:
\begin{equation}
    \mathcal{L}
      = \mathcal{L}_{\mathrm{CE}}
      + \lambda_{\mathrm{det}}\,\mathcal{L}_{\mathrm{det}},
    \qquad \lambda_{\mathrm{det}} = 0.1.
    \label{eq:loss}
\end{equation}
Because the detector acts on discrete decoded text,
$\mathcal{L}_{\mathrm{det}}$ is not directly differentiable;
we optimize it with the REINFORCE
estimator~\citep{williams1992simple}, sampling
$\hat{x}\sim p_\theta$, scoring it with the detector, and
subtracting a moving-average baseline~$b$ to reduce variance:
\begin{equation}
    \nabla_\theta \mathcal{L}_{\mathrm{det}}
      = \mathbb{E}_{\hat{x}\sim p_\theta}\!\bigl[
        (\pai(\hat{x}) - b)\,
        \nabla_\theta \log p_\theta(\hat{x})
        \bigr],
    \label{eq:reinforce}
\end{equation}
where $\pai(\hat{x})$ is the detector's confidence that the
generated text~$\hat{x}$ is AI-written.
Rather than simply maximizing evasion, this signal acts as a
quality regularizer: it encourages the model to achieve
evasion through targeted stylistic shifts rather than content
distortion, yielding lower perplexity at comparable evasion
rates (see A2 ablation, \S\ref{sec:ablation}).

%% ────────────────────────────────────────────────────
\subsection{SDEdit Inference}
\label{sec:inference}

At test time, we adapt the SDEdit paradigm~\citep{meng2022sdedit}
to token embeddings.
We first embed the input AI text to obtain
$\mathbf{e}_{\mathrm{AI}}$ and compute its Qwen representations
$\mathbf{H}_{\mathrm{qwen}}$ (\S\ref{sec:conditioning}).
We then corrupt the embeddings at noise level $\gamma_{\mathrm{start}}$:
$\mathbf{z}_0
  = \alpha(\gamma_{\mathrm{start}})\,\mathbf{e}_{\mathrm{AI}}
  + \sigma(\gamma_{\mathrm{start}})\,\boldsymbol{\epsilon}$.
Starting from $\mathbf{z}_0$, we perform $T{=}64$ Euler steps
conditioned on $\mathbf{H}_{\mathrm{qwen}}$ through the
cross-attention adapters, then decode via $\arg\max$.
The parameter $\gamma_{\mathrm{start}}$ serves as a
continuous diagnostic axis: low~$\gamma$ yields high
similarity and low evasion; high~$\gamma$ applies stronger
transformation at a modest cost in semantic similarity.

%% ────────────────────────────────────────────────────
\subsection{RateAudit: Document-Level Scheduling}
\label{sec:allocator}

Deployed AIGC detectors typically process documents in
fixed-length windows ($\leq$512 tokens), then aggregate
window-level scores into a document verdict.
RateAudit exploits this aggregation step to
demonstrate that pre-specified target detection rates can be
achieved on long texts while preserving semantic
coherence---directly questioning the reliability of
detection-rate-based document verdicts.

Given a document~$D$ and a target rate~$r_{\mathrm{target}}$,
we divide~$D$ into chunks $\{c_1,\ldots,c_n\}$ of at most 512 tokens.
We score each chunk and compute the document-level score as
$\bar{s}=\sum_i w_i\,\pai(c_i)$, where $w_i{=}|c_i|/|D|$.
We then greedily rewrite the chunk with the highest score,
using an adaptive~$\gamma$ schedule ($6.0{\to}6.5{\to}7.0{\to}7.5$),
re-scoring after each pass.
The procedure terminates when
$|\bar{s}-r_{\mathrm{target}}|\leq\epsilon$, with $\epsilon{=}0.02$.
The key empirical insight is that most documents have a small
number of high-$\pai$ chunks; rewriting only these
suffices to shift the aggregate score while largely preserving
the document's content.

\section{Experiments}
\label{sec:experiments}

\begin{table*}[!t]
\centering
\small
\begin{tabular}{l c cccc c}
\toprule
Method & $\gamma$ & $\pai\,{\downarrow}$ & Evade@0.5$\uparrow$ & Evade@0.3$\uparrow$ & Sem.Sim$\uparrow$ & PPL \\
\midrule
Original AI text & -- & 0.985 & 0.0\% & 0.0\% & 1.000 & 10.7 \\
Human text & -- & 0.015 & 100.0\% & 100.0\% & -- & 16.5 \\
\midrule
Synonym Substitution & -- & 0.985 & 0.0\% & 0.0\% & 1.000 & 11.0 \\
Backtranslation & -- & 0.203 & 82.6\% & 76.8\% & 0.852 & 14.6 \\
LLM Rewrite & -- & 0.993 & 0.3\% & 0.1\% & 0.964 & 9.5 \\
\midrule
\multirow{5}{*}{\styleshield{} (Ours)} 
& 5.0 & 0.703 & 30.1\% & 25.1\% & 0.948 & 30.1 \\
& 5.5 & 0.535 & 36.8\% & 31.3\% & 0.947 & 22.5 \\
& 6.0 & 0.301 & 52.6\% & 46.8\% & 0.942 & 20.3 \\
& 6.5 & 0.124 & 79.7\% & 72.5\% & 0.935 & 21.1 \\
& 7.0 & 0.072 & 94.6\% & 88.7\% & 0.928 & 21.2 \\
\bottomrule
\end{tabular}
\caption{Main results on the training detector (Det-v3). \styleshield{} achieves continuous control over the evasion--similarity trade-off via $\gamma$. At $\gamma{=}7.0$, it achieves 94.6\% evasion while maintaining 0.928 semantic similarity. Baselines offer no such controllability: Backtranslation achieves moderate evasion but at severe semantic cost (0.852); LLM Rewrite and Synonym Substitution fail to evade. PPL of human text (${\sim}$16.5) and AI text (${\sim}$10.7) serve as references; \styleshield{}'s PPL (${\sim}$21) exceeds the human baseline, reflecting successful injection of stylistic variability beyond the narrow AI distribution.}
\label{tab:main_results}
\end{table*}

\subsection{Experimental Setup}
\label{sec:setup}

We use the 436K multi-domain parallel corpus described in \S\ref{sec:training} (346K social media, 68K news, 22K academic), holding out 1{,}000 balanced samples for evaluation.
All texts are truncated to 512 tokens.

On this test set, we evaluate against four AIGC detectors:
Det-v3 (training detector, fine-tuned Chinese BERT),
Det-v2 (earlier version of the same family),
ANX-BERT (public Chinese AI text detector), and
GPT2-Det (GPT-2-based, multilingual).
Only Det-v3 is used during training; the other three are held out to test cross-detector generalization.
For comparison, we include three adversarial probing baselines:
Synonym Substitution (Chinese synonym dictionary, structure-preserving),
Backtranslation (Chinese$\to$English$\to$Chinese via NLLB-200~\citep{fan2022nllb}), and
LLM Rewrite (Qwen2.5-7B-Instruct prompted to humanize).

We evaluate detection evasion and text quality.
For evasion, we report $\pai$, the mean detector confidence that the text is AI-generated, and Evade@0.5 and Evade@0.3, the percentages of samples scoring below the respective detection thresholds.
Lower $\pai$ and higher evasion rates indicate more successful evasion.
We assess semantic preservation using the cosine similarity of Qwen embeddings between input and output, and report perplexity (PPL) measured by a Chinese GPT-2 model trained for neither detection nor style transfer.
For reference, original AI text has PPL${\approx}$10.7 and human text has PPL${\approx}$16.5.

The trainable model totals ${\sim}$113M parameters (${\sim}$85M backbone + ${\sim}$28M cross-attention adapters); the Qwen2.5-7B-Instruct encoder contributes an additional 7B frozen parameters.
Inputs are tokenized with the Chinese BERT tokenizer (bert-base-chinese; ${\sim}$21K vocabulary), max length 512.
Training uses 16$\times$A800 GPUs (PyTorch DDP, global batch size 1{,}024), AdamW~\citep{loshchilov2019decoupled} ($\beta_1{=}0.9$, $\beta_2{=}0.999$, weight decay $0.01$), with learning rates $10^{-4}$ (backbone) and $5{\times}10^{-4}$ (new parameters), linear warmup over 1{,}000 steps and cosine decay.
The main model trains for 100K steps (${\sim}$48h); ablation variants for 60K steps (${\sim}$29h).
Inference uses $T{=}64$ Euler steps on a single A800; each sample takes ${\sim}$6s, with the full 1{,}000-sample suite completing in ${\sim}$1.7h per $\gamma$.

\subsection{Main Results}
\label{sec:main_results}

Table~\ref{tab:main_results} presents the main comparison on the training detector (Det-v3).
\styleshield{} with $\gamma{=}7.0$ achieves the best overall balance: 94.6\% evasion at the 0.5 threshold, 88.7\% at the stricter 0.3 threshold, with 0.928 semantic similarity.
At the most aggressive setting ($\gamma{=}7.0$), the mean $\pai$ drops to 0.072---a 13.7$\times$ reduction from the Synonym baseline (0.985) and far below the LLM Rewrite baseline (0.993).

Backtranslation achieves a competitive evasion rate (82.6\%) but at a severe cost to semantic similarity (0.852 vs.\ 0.928), and its PPL (14.6) falls below the human reference (16.5), suggesting the output has acquired an unnatural regularity distinct from genuine human writing.
Despite preserving semantic similarity, the LLM Rewrite baseline achieves only 0.3\% evasion against Det-v3 under our evaluation setting.

A distinguishing feature of \styleshield{} is continuous controllability: varying $\gamma$ from 5.0 to 7.0 traces a smooth Pareto frontier between semantic preservation (0.948 at $\gamma{=}5.0$) and evasion strength (94.6\% at $\gamma{=}7.0$)---a continuous diagnostic axis unavailable to any baseline.
Human evaluation shows a corresponding trade-off between perceived humanness and content preservation (Table~\ref{tab:human_eval}).
Four native Chinese speakers with NLP backgrounds rated outputs for 100 sampled inputs at $\gamma\!\in\!\{6.5,7.0\}$ on a 1--5 scale, blind to $\gamma$.
Increasing $\gamma$ from 6.5 to 7.0 raises perceived humanness but lowers content preservation and overall quality.
Details are provided in Appendix~\ref{sec:app_human_eval}.

\begin{table}[H]
\centering
\small
\begin{tabular}{l cccc}
\toprule
$\gamma$ & Flu. & Cont. & Hum. & Over. \\
\midrule
6.5 & 4.02$_{\pm0.68}$ & 4.08$_{\pm0.71}$ & 3.54$_{\pm0.78}$ & 3.89$_{\pm0.67}$ \\
7.0 & 3.81$_{\pm0.76}$ & 3.68$_{\pm0.73}$ & 3.96$_{\pm0.69}$ & 3.72$_{\pm0.68}$ \\
\bottomrule
\end{tabular}
\caption{Human evaluation (mean $\pm$ std, 1--5 scale; $n{=}100$ per $\gamma$), averaged across four annotators. Flu.: fluency; Cont.: content preservation; Hum.: humanness; Over.: overall quality.}
\label{tab:human_eval}
\end{table}

\begin{table*}[!t]
\centering
\small
\begin{tabular}{l cccc cccc}
\toprule
& \multicolumn{4}{c}{$\pai\,{\downarrow}$} & \multicolumn{4}{c}{Evade@0.5$\uparrow$} \\
\cmidrule(lr){2-5} \cmidrule(lr){6-9}
Method & Det-v3$^*$ & Det-v2 & ANX & GPT2 & Det-v3$^*$ & Det-v2 & ANX & GPT2 \\
\midrule
Synonym Sub. & 0.985 & 0.424 & 0.856 & 0.153 & 0.0\% & 58.2\% & 12.2\% & 84.9\% \\
Backtrans. & 0.203 & 0.140 & 0.270 & 0.003 & 82.6\% & 88.4\% & 74.4\% & 99.8\% \\
LLM Rewrite & 0.993 & 0.803 & 0.904 & 0.245 & 0.3\% & 18.0\% & 7.2\% & 76.6\% \\
\midrule
\styleshield{} ($\gamma{=}6.0$) & 0.301 & 0.036 & 0.100 & 0.014 & 52.6\% & 97.1\% & 92.5\% & 98.8\% \\
\styleshield{} ($\gamma{=}6.5$) & 0.124 & 0.007 & 0.039 & 0.013 & 79.7\% & 99.7\% & 98.6\% & 98.7\% \\
\styleshield{} ($\gamma{=}7.0$) & 0.072 & 0.002 & 0.035 & 0.010 & 94.6\% & 100.0\% & 99.0\% & 99.1\% \\
\bottomrule
\end{tabular}
\caption{Cross-detector generalization. Only Det-v3 (marked $^*$) is used during training; the other three detectors are unseen. \styleshield{} achieves $\geq$99\% evasion on all unseen detectors at $\gamma{=}7.0$, consistent with the hypothesis that the style transfer learns genuine distributional shifts rather than detector-specific adversarial features. No baseline achieves consistent evasion across all detectors.}
\label{tab:cross_detector}
\end{table*}

\subsection{Cross-Detector Generalization}
\label{sec:cross_detector}

\begin{figure}[!t]
    \centering
    \includegraphics[width=\columnwidth]{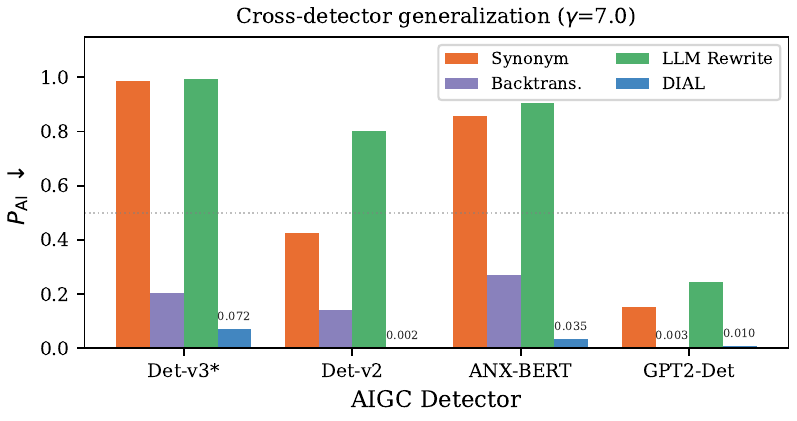}
    \caption{Cross-detector $\pai$ at $\gamma{=}7.0$. \styleshield{} achieves near-zero $\pai$ on all detectors, while baselines show inconsistent performance.}
    \label{fig:cross_detector}
\end{figure}

A critical question is whether \styleshield{}, trained with Det-v3 feedback, generalizes to unseen detectors.
Table~\ref{tab:cross_detector} and Figure~\ref{fig:cross_detector} show results across all four detectors at $\gamma{=}7.0$.

\styleshield{} achieves near-perfect evasion on all three unseen detectors: 100.0\% on Det-v2, 99.0\% on ANX-BERT, and 99.1\% on GPT2-Det.
This strong generalization is consistent with the hypothesis that the style transfer learns genuine distributional shifts rather than detector-specific adversarial features.
In contrast, baselines show highly inconsistent cross-detector performance: Synonym Substitution achieves 84.9\% on GPT2-Det but only 0.0\% on Det-v3; Backtranslation is strong on GPT2-Det (99.8\%) but weaker on ANX-BERT (74.4\%).

\begin{table*}[!t]
\centering
\small
\begin{tabular}{l c cccc c}
\toprule
Model Variant & $\gamma$ & $\pai\,{\downarrow}$ & Evade@0.5$\uparrow$ & Evade@0.3$\uparrow$ & Sem.Sim$\uparrow$ & PPL \\
\midrule
\multirow{2}{*}{Full Model (Layer 14)} 
& 6.5 & 0.124 & 79.7\% & 72.5\% & 0.935 & 21.1 \\
& 7.0 & 0.072 & 94.6\% & 88.7\% & 0.928 & 21.2 \\
\midrule
\multirow{2}{*}{A1: Single-Domain Only}
& 6.5 & 0.138 & 77.2\% & 69.8\% & 0.931 & 22.4 \\
& 7.0 & 0.081 & 92.1\% & 85.3\% & 0.923 & 22.5 \\
\midrule
\multirow{2}{*}{A2: w/o Detector Reward}
& 6.5 & 0.145 & 89.4\% & 85.5\% & 0.928 & 26.4 \\
& 7.0 & 0.086 & 92.4\% & 84.4\% & 0.921 & 26.1 \\
\midrule
\multirow{2}{*}{A3: Split Layer 7}
& 6.5 & 0.014 & 100.0\% & 100.0\% & 0.904 & 99.9 \\
& 7.0 & 0.011 & 100.0\% & 100.0\% & 0.900 & 106.9 \\
\midrule
\multirow{2}{*}{A4: Split Layer 21}
& 6.5 & 0.019 & 99.8\% & 99.4\% & 0.872 & 50.5 \\
& 7.0 & 0.019 & 99.7\% & 99.5\% & 0.858 & 49.1 \\
\midrule
\multirow{2}{*}{A5: w/o Qwen Cond.}
& 6.5 & 0.021 & 99.6\% & 99.1\% & 0.747 & 118.3 \\
& 7.0 & 0.016 & 99.9\% & 99.7\% & 0.731 & 132.7 \\
\bottomrule
\end{tabular}
\caption{Ablation study on Det-v3. A1: Single-domain training slightly degrades evasion and similarity. A2: Removing the detector reward increases PPL and lowers semantic similarity at both evaluated strengths. A3/A4: Among the three tested Qwen split depths, Layer~14 provides the best balance between evasion and semantic preservation. A5: Removing Qwen conditioning entirely achieves near-perfect evasion but catastrophic quality degradation (PPL$>$118, Sim$<$0.75), confirming that semantic conditioning is essential for meaningful style transfer.}
\label{tab:ablation}
\end{table*}

\subsection{Ablation Study}
\label{sec:ablation}

\begin{figure}[!t]
    \centering
    \includegraphics[width=\columnwidth]{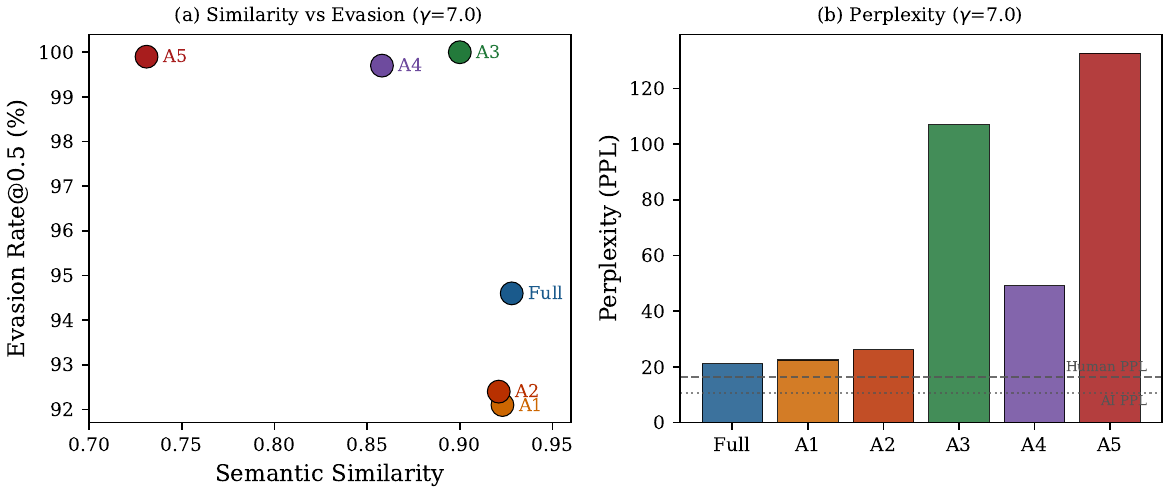}
    \caption{Ablation analysis at $\gamma{=}7.0$. (a)~Similarity vs.\ evasion: the full model achieves the best quality-evasion balance. (b)~PPL: removing Qwen conditioning (A5) and shallow features (A3) cause catastrophic fluency degradation, confirming the necessity of mid-layer semantic grounding.}
    \label{fig:ablation}
\end{figure}

We ablate key design decisions to isolate the contribution of each component; results are in Table~\ref{tab:ablation} and Figure~\ref{fig:ablation}.

\noindent A1: Single-Domain Training (Social Media Only).\quad
Training on only social media data (346K pairs, without news and academic domains) tests the contribution of multi-domain diversity.
A1 achieves slightly lower evasion (92.1\% vs.\ 94.6\% at $\gamma{=}7.0$) and worse semantic similarity (0.923 vs.\ 0.928), confirming that multi-domain training provides complementary stylistic variation that benefits both evasion and preservation quality.

\noindent A2: Without Detector Reward ($\lambda_{\mathrm{det}}{=}0$).\quad
Removing the detector reward yields comparable evasion rates (92.4\% vs.\ 94.6\% at $\gamma{=}7.0$ on Det-v3) but at the cost of lower semantic similarity (0.921 vs.\ 0.928) and substantially higher PPL (26.1 vs.\ 21.2).
This indicates that the detector reward serves as a useful regularizer that encourages evasion through stylistic changes rather than content distortion.

\noindent A3/A4: Qwen Split Layer Depth (Layer 7 vs.\ 14 vs.\ 21).\quad
Both shallower (L=7) and deeper (L=21) features achieve near-perfect evasion ($\geq$99.7\%) but at substantial quality cost: L=7 yields catastrophic PPL (99.9--106.9), whereas L=21 has lower PPL (49.1--50.5) but the lowest semantic similarity among the three tested depths (0.858--0.872).
Among these depths, Layer~14 provides the best balance between evasion and semantic preservation.

\noindent A5: Without Qwen Conditioning.\quad
Removing Qwen cross-attention conditioning entirely isolates the contribution of semantic grounding.
Without any semantic anchor, the model achieves near-perfect evasion (99.9\% at $\gamma{=}7.0$) but at catastrophic quality cost: PPL explodes to 132.7 (vs.\ 21.2 for the full model) and semantic similarity plummets to 0.731 (vs.\ 0.928).
This confirms that Qwen conditioning is the critical component for meaningful style transfer---without it, the model degenerates into unconstrained text perturbation that destroys content while trivially fooling detectors.

\subsection{RateAudit: Long-Document Results}
\label{sec:allocator_results}

Table~\ref{tab:allocator} reports RateAudit results averaged over 50 documents (8{,}000--12{,}000 characters each) across six target rates (10\%--60\%).
RateAudit achieves precise control with mean deviation $\leq$1.4pp from target, while largely preserving the document's content.
This directly questions the evidentiary value of percentage-based detection verdicts: if the reported rate can be reliably shifted to an arbitrary target while preserving semantic coherence, such verdicts carry no diagnostic weight.

\begin{table}[H]
\centering
\small
\begin{tabular}{c cc cc}
\toprule
Target & Achieved & Chunks & Rewrite & Time \\
Rate & $\pai$(\%) & Rewritten & Ratio & (s/doc) \\
\midrule
10\% & 10.1$\pm$0.8 & 19.6 / 23 & 88.7\% & 271$\pm$18 \\
20\% & 21.3$\pm$1.2 & 16.8 / 23 & 75.4\% & 232$\pm$21 \\
30\% & 30.5$\pm$1.4 & 14.7 / 23 & 66.1\% & 204$\pm$19 \\
40\% & 40.8$\pm$1.1 & 12.9 / 23 & 57.8\% & 179$\pm$16 \\
50\% & 50.2$\pm$0.9 & 10.8 / 23 & 48.6\% & 150$\pm$14 \\
60\% & 60.7$\pm$1.3 &  8.2 / 23 & 37.1\% & 112$\pm$12 \\
\bottomrule
\end{tabular}
\caption{RateAudit on Det-v3: 50 documents (8{,}000--12{,}000 characters; mean 23 chunks). Original documents have mean $\pai{=}98.7\%$. Runtime is measured on one A800 GPU.}
\label{tab:allocator}
\end{table}

\section{Analysis}
\label{sec:analysis}

\subsection{Controllability: The $\gamma$ Curve}

\begin{figure}[!t]
    \centering
    \includegraphics[width=\columnwidth]{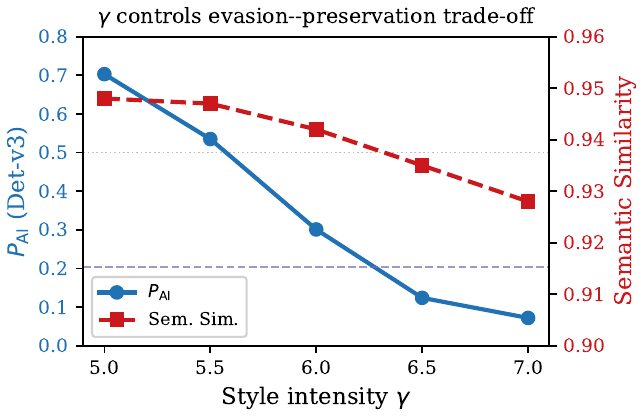}
    \caption{$\pai$ (solid) and semantic similarity (dashed) as functions of $\gamma$ on Det-v3. The backtranslation baseline's $\pai$ is shown as a dashed reference line. \styleshield{} provides smooth, monotonic control; the ``sweet spot'' lies around $\gamma{=}6.5$--$7.0$.}
    \label{fig:gamma_curve}
\end{figure}

\begin{figure}[!t]
    \centering
    \includegraphics[width=\columnwidth]{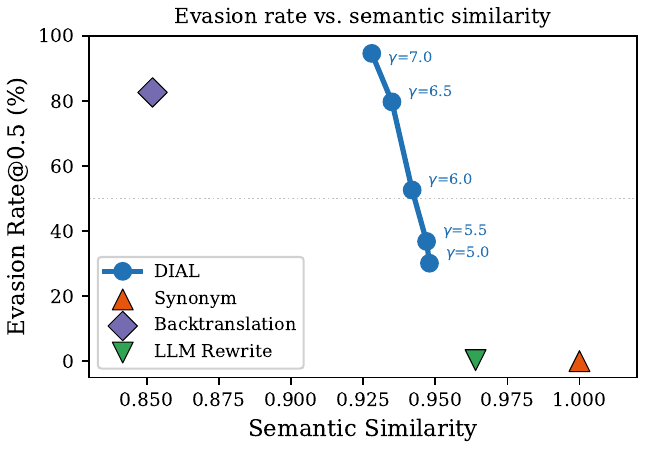}
    \caption{Evasion--similarity trade-off across $\gamma$. \styleshield{} traces a controllable curve and, at $\gamma{=}7.0$, outperforms Backtranslation in both evasion rate and semantic similarity.}
    \label{fig:pareto}
\end{figure}

Figure~\ref{fig:gamma_curve} plots key metrics as a function of $\gamma$.
As $\gamma$ increases from 5.0 to 7.0, $\pai$ on Det-v3 decreases monotonically from 0.703 to 0.072 while semantic similarity drops only from 0.948 to 0.928, confirming that $\gamma$ provides reliable continuous control.
At comparable evasion rates (79.7\% vs.\ 82.6\%), \styleshield{} achieves higher semantic similarity than Backtranslation (0.935 vs.\ 0.852; Figure~\ref{fig:pareto}).
On unseen detectors the curve is even more favorable---at $\gamma{=}6.0$, evasion already exceeds 92\% on all three held-out detectors while maintaining 0.942 similarity, indicating that moderate transformation suffices to expose cross-detector vulnerabilities.

\subsection{Perplexity Analysis}

An initially counterintuitive finding is that \styleshield{}'s output PPL (${\sim}$21) exceeds that of human text (${\sim}$16.5), which in turn exceeds original AI text (${\sim}$10.7).
We argue this pattern is desirable: low PPL is precisely the statistical regularity that detectors exploit.
AI-generated text exhibits an unnaturally low perplexity---it follows high-probability paths through the language model's distribution, producing statistically regular but stylistically flat output.
Human text, by contrast, contains idiosyncratic word choices, colloquialisms, and stylistic variation that increase perplexity.
\styleshield{}'s elevated PPL reflects successful injection of human-like variability, pushing the text's statistical profile beyond the narrow AI distribution.

The ablation with shallow Qwen features (A3, Layer 7) exhibits extreme PPL (99.9--106.9), and removing Qwen conditioning entirely (A5) pushes PPL further to 118--133, indicating that without strong semantic grounding, the model introduces too much variation, degrading fluency.
The full model with layer-14 features achieves the right balance: enough variation to evade detection (PPL$\approx$21) while remaining within the range of fluent text.

\subsection{What Does the Detector Reward Actually Do?}

The A2 ablation (no detector reward) achieves comparable evasion to the full model (e.g., 92.4\% vs.\ 94.6\% at $\gamma{=}7.0$), suggesting the detector reward's primary contribution is not to evasion itself.
However, the quality metrics tell a different story: without the reward, PPL increases to 26.1--26.4 (vs.\ 21.1--21.2), and semantic similarity drops to 0.921 (vs.\ 0.928).
These results indicate that the reward helps the model achieve distributional shifts with less damage to text quality.

\subsection{Qualitative Examples}

As $\gamma$ increases, formulaic AI template language is progressively replaced by colloquial phrasing while core content is preserved, and $\pai$ declines smoothly across all domains---confirming that $\gamma$ functions as a fine-grained diagnostic axis rather than a binary switch.
We provide three representative input--output pairs (one per domain) at $\gamma\in\{6.0,6.5,7.0\}$, with English glosses for non-Chinese-speaking reviewers, in Figure~\ref{fig:qualitative} (Appendix~\ref{sec:app_qualitative}).

\section{Conclusion}
\label{sec:conclusion}

We presented \styleshield{}, a source-conditioned flow-matching framework for continuously controllable text style transfer, exposing the fragility of AIGC detectors through stylistic transformation in embedding space.
By combining a DiT backbone with zero-initialized cross-attention adapters conditioned on frozen Qwen2.5-7B representations, \styleshield{} at $\gamma{=}7.0$ achieves 94.6\% evasion on the training detector and $\geq$99\% on three unseen detectors, while maintaining 0.928 semantic similarity.
RateAudit further demonstrates that document-level detection verdicts can be shifted toward pre-specified target rates---a result that, in our view, argues against treating percentage-based scores as the sole evidence in consequential decisions.

\paragraph{Broader Impact.}
We build \styleshield{} not to facilitate evasion but to provide an independent audit: a single lightweight model is enough to evade the four detectors we evaluate, suggesting that origin-based text classification, at least as currently deployed, is fragile for high-stakes decisions.
We urge the community to move toward process-based and quality-centered evaluation (see Ethics Statement for detailed recommendations).

\section*{Limitations}
Our approach currently focuses on Chinese text, and its generalization to other languages remains to be validated.
Our paired corpus uses Qwen2.5-7B-Instruct to generate the AI texts and Det-v3 to filter the pairs. Generalization to other generators and unfiltered text distributions remains to be evaluated.
The system relies on access to a target AIGC detector during training (though it generalizes to unseen detectors at test time), which may not always be available.
Additionally, our evaluation of text quality relies primarily on automatic metrics; while perplexity and semantic similarity provide useful proxies, they do not fully capture the nuanced aspects of naturalness that human readers perceive.

\section*{Ethics Statement}
\label{sec:ethics}

\styleshield{} is built as a diagnostic probe, not an evasion toolkit.
We release it to make visible a reliability problem in AIGC detection that already affects real people, and to argue that classifying text by its origin is the wrong basis for decisions with serious consequences.

\noindent The Cost Asymmetry.\quad
Table~\ref{tab:cost} normalizes AIGC detection fees and mainstream LLM API prices to a common unit (USD per million output tokens).
The gap is staggering: CNKI charges \$641/M\,tok---6{,}410$\times$ the rate of the cheapest commercial LLM (\$0.10/M\,tok), and 26$\times$ more than even the costliest frontier models (${\sim}$\$25/M\,tok).
Even the cheapest detection platform (PaperPass, \$321/M\,tok) costs 3{,}210$\times$ more.
Put differently, a single detection check costs more per token than running the most capable commercial LLM available today by an order of magnitude.
This pricing structure creates a profitable market built on classifiers whose reliability has never been independently audited: CNKI's digital publishing arm reported 49.7\% gross margins and 148 million CNY net profit in H1\,2024 alone.\footnote{Source: Tongfang Co., Ltd.\ (SSE: 600100) 2024 semi-annual report, CNKI digital publishing segment.}
We believe the research community deserves to know what this expenditure actually buys.

\begin{table}[!t]
\centering
\small
\setlength{\tabcolsep}{4pt}
\begin{tabular}{l l r r}
\toprule
Category & Service / Model & \$/M tok & Ratio \\
\midrule
\multirow{4}{*}{\rotatebox{90}{Detection}}
& CNKI & 641$^\dagger$ & 6{,}410$\times$ \\
& Weipu & 427$^\dagger$ & 4{,}270$\times$ \\
& Wanfang & 427$^\dagger$ & 4{,}270$\times$ \\
& PaperPass & 321$^\dagger$ & 3{,}210$\times$ \\
\midrule
\multirow{10}{*}{\rotatebox{90}{Commercial LLM API}}
& Claude Opus 4.6 & 25.00 & 250$\times$ \\
& Claude Sonnet 4.6 & 15.00 & 150$\times$ \\
& Claude Haiku 4.5 & 5.00 & 50$\times$ \\
& GPT-4o & 10.00 & 100$\times$ \\
& GPT-4.1 & 2.00 & 20$\times$ \\
& GPT-4.1-mini & 0.50 & 5.0$\times$ \\
& Kimi K2 & 3.00 & 30$\times$ \\
& Qwen-Max & 0.83$^\ddagger$ & 8.3$\times$ \\
& Qwen-Turbo & 0.10$^\ddagger$ & 1.0$\times$ \\
& DeepSeek-V3 & 0.28 & 2.8$\times$ \\
\bottomrule
\end{tabular}
\caption{AIGC detection vs.\ LLM API costs (USD/M output tokens).
Ratio: relative to the cheapest LLM (Qwen-Turbo).
$^\dagger$Converted from per-character rates (1K Chinese chars\,$\approx$\,650 tokens; 1\,USD\,$=$\,7.2\,CNY).
$^\ddagger$Converted from Alibaba Cloud CNY rates.
Prices are non-batch, non-cached rates; the cost asymmetry reported here reflects the market structure as of the submission date.}
\label{tab:cost}
\end{table}

\noindent False Positives Cause Real Harm.\quad
Reported false-positive rates for mainstream AIGC detectors range from 5\% to 12\%.
In a widely covered 2025 incident, a Renmin University professor's three-year ethnographic fieldwork was flagged as highly suspected AI-generated~\citep{sun2025aigc}.
Classical Chinese texts such as Wang Bo's Preface to the Pavilion of Prince Teng (675\,CE) have been scored above 50\% AI-generated by commercial detectors~\citep{sun2025aigc}.
Students whose theses exceed a 20\% AI-detection threshold face degree revocation; faculty face suspension of supervisory privileges~\citep{restofworld2025aigc}.
When the penalty for a false positive is career-altering and the classifier's error rate is in the single digits, the expected harm is not a corner case---it is a systemic risk.

\noindent Dual-Use Considerations.\quad
We acknowledge that \styleshield{} could, in principle, be misused to disguise AI-generated text.
However, the techniques it employs---flow-matching style transfer, semantic conditioning, iterative denoising---are well within the capability of any moderately resourced actor; withholding a research implementation does not raise the barrier to misuse.
What does raise the barrier to accountability is the continued deployment of detectors whose fragility has not been publicly documented.
Our contribution is to provide that documentation.
Accordingly, we frame \styleshield{} as an auditing instrument rather than a deployable evasion service, and we advocate responsible, research-oriented use.

\noindent Recommended Policy Direction.\quad
Rather than an arms race between generators and detectors, we advocate for:
(1)~process-based assessment: evaluating drafts, revision logs, and oral defenses rather than a single classifier score;
(2)~mandatory error-rate disclosure: requiring detection vendors to publish independently audited false-positive and false-negative rates before their tools can be used in consequential decisions;
and (3)~quality-centered evaluation: judging text by the understanding, reasoning, and originality it demonstrates, regardless of how it was produced.

\bibliography{references}
  
\appendix
\section{Extended Ablation Results}
\label{sec:app_ablation}

Table~\ref{tab:ablation_full} presents the full ablation results across all $\gamma$ values and all four detectors.

\begin{table*}[!t]
\centering
\small
% Local row spacing keeps the dense appendix table readable and avoids stretched section gaps below it.
\renewcommand{\arraystretch}{1.15}
\begin{tabular}{l c cccc cc}
\toprule
Variant & $\gamma$ & Det-v3$^*$ & Det-v2 & ANX & GPT2 & Sim & PPL \\
\midrule
\multirow{5}{*}{Full Model}
& 5.0 & 0.703 & 0.150 & 0.367 & 0.031 & 0.948 & 30.1 \\
& 5.5 & 0.535 & 0.104 & 0.247 & 0.030 & 0.947 & 22.5 \\
& 6.0 & 0.301 & 0.036 & 0.100 & 0.014 & 0.942 & 20.3 \\
& 6.5 & 0.124 & 0.007 & 0.039 & 0.013 & 0.935 & 21.1 \\
& 7.0 & 0.072 & 0.002 & 0.035 & 0.010 & 0.928 & 21.2 \\
\midrule
\multirow{5}{*}{A1: Single-Domain}
& 5.0 & 0.718 & 0.163 & 0.389 & 0.035 & 0.944 & 31.5 \\
& 5.5 & 0.561 & 0.112 & 0.268 & 0.033 & 0.942 & 23.8 \\
& 6.0 & 0.327 & 0.041 & 0.118 & 0.017 & 0.937 & 21.6 \\
& 6.5 & 0.138 & 0.009 & 0.047 & 0.015 & 0.931 & 22.4 \\
& 7.0 & 0.081 & 0.003 & 0.041 & 0.012 & 0.923 & 22.5 \\
\midrule
\multirow{5}{*}{A2: w/o Det.~Reward}
& 5.0 & 0.710 & 0.133 & 0.363 & 0.036 & 0.946 & 33.1 \\
& 5.5 & 0.593 & 0.048 & 0.181 & 0.027 & 0.945 & 26.5 \\
& 6.0 & 0.329 & 0.005 & 0.060 & 0.019 & 0.937 & 26.0 \\
& 6.5 & 0.145 & 0.001 & 0.031 & 0.010 & 0.928 & 26.4 \\
& 7.0 & 0.086 & 0.001 & 0.027 & 0.011 & 0.921 & 26.1 \\
\midrule
\multirow{5}{*}{A3: Split Layer 7}
& 5.0 & 0.570 & 0.054 & 0.222 & 0.025 & 0.943 & 44.5 \\
& 5.5 & 0.268 & 0.002 & 0.061 & 0.010 & 0.932 & 53.9 \\
& 6.0 & 0.043 & 0.000 & 0.032 & 0.005 & 0.916 & 79.0 \\
& 6.5 & 0.014 & 0.000 & 0.028 & 0.004 & 0.904 & 99.9 \\
& 7.0 & 0.011 & 0.000 & 0.026 & 0.002 & 0.900 & 106.9 \\
\midrule
\multirow{5}{*}{A4: Split Layer 21}
& 5.0 & 0.590 & 0.037 & 0.234 & 0.030 & 0.936 & 49.2 \\
& 5.5 & 0.249 & 0.002 & 0.067 & 0.016 & 0.917 & 47.7 \\
& 6.0 & 0.044 & 0.000 & 0.033 & 0.004 & 0.892 & 50.2 \\
& 6.5 & 0.019 & 0.000 & 0.032 & 0.003 & 0.872 & 50.5 \\
& 7.0 & 0.019 & 0.000 & 0.028 & 0.003 & 0.858 & 49.1 \\
\midrule
\multirow{5}{*}{A5: w/o Qwen Cond.}
& 5.0 & 0.412 & 0.028 & 0.187 & 0.022 & 0.791 & 85.4 \\
& 5.5 & 0.138 & 0.003 & 0.058 & 0.009 & 0.772 & 97.6 \\
& 6.0 & 0.041 & 0.000 & 0.034 & 0.005 & 0.758 & 110.2 \\
& 6.5 & 0.021 & 0.000 & 0.030 & 0.003 & 0.747 & 118.3 \\
& 7.0 & 0.016 & 0.000 & 0.027 & 0.002 & 0.731 & 132.7 \\
\bottomrule
\end{tabular}
\caption{Full ablation results ($\pai$ values) across all detectors and $\gamma$ values. Det-v3 (marked $^*$) is the training detector.}
\label{tab:ablation_full}
\end{table*}

\section{Human Evaluation}
\label{sec:app_human_eval}

To complement automatic metrics, we conduct a human evaluation on 100 randomly sampled test instances (34 news, 33 social media, 33 academic).
Four native Chinese speakers with NLP backgrounds independently rated each \styleshield{} output at $\gamma\!\in\!\{6.5, 7.0\}$ on four dimensions using a predefined 1--5 Likert rubric:
Fluency (linguistic naturalness),
Content (preservation of source information),
Humanness (perceived human authorship), and
Overall quality.
Annotation was performed blind to $\gamma$ values.
Results are in Table~\ref{tab:human_eval}.

At $\gamma{=}6.5$, outputs receive high marks for fluency (4.02) and content preservation (4.08), confirming that moderate transformation retains both readability and fidelity.
Humanness is relatively lower (3.54), consistent with residual AI-like phrasing at this intensity.
At $\gamma{=}7.0$, humanness rises to 3.96---the highest single-dimension score---indicating that stronger transformation successfully removes detectable AI patterns, though content preservation decreases to 3.68.
This trade-off mirrors the automatic $\pai$--similarity curve (\S\ref{sec:analysis}) and confirms that the continuous diagnostic axis is perceptible to human raters.

\section{RateAudit Details}
\label{sec:app_allocator}

RateAudit operates as a greedy scheduler.
Given a document $D$, a target detection rate $r_{\text{target}}$, and a tolerance $\epsilon$, it first splits $D$ into chunks $\{c_1,\ldots,c_n\}$ and scores each chunk with the detector to obtain $s_i = \pai(c_i)$.
At each iteration, it selects the chunk with the highest $s_i$, rewrites it with \styleshield{} at the smallest $\gamma$ that reduces its score, and updates the length-weighted document average $\bar{s}=\sum_i w_i s_i$.
The loop terminates when $|\bar{s} - r_{\text{target}}| \leq \epsilon$.
This selective rewriting shifts the document-level score while largely preserving the document's content.

\section{Qualitative Examples}
\label{sec:app_qualitative}

\begin{figure*}[!t]
    \centering
    \includegraphics[width=\textwidth]{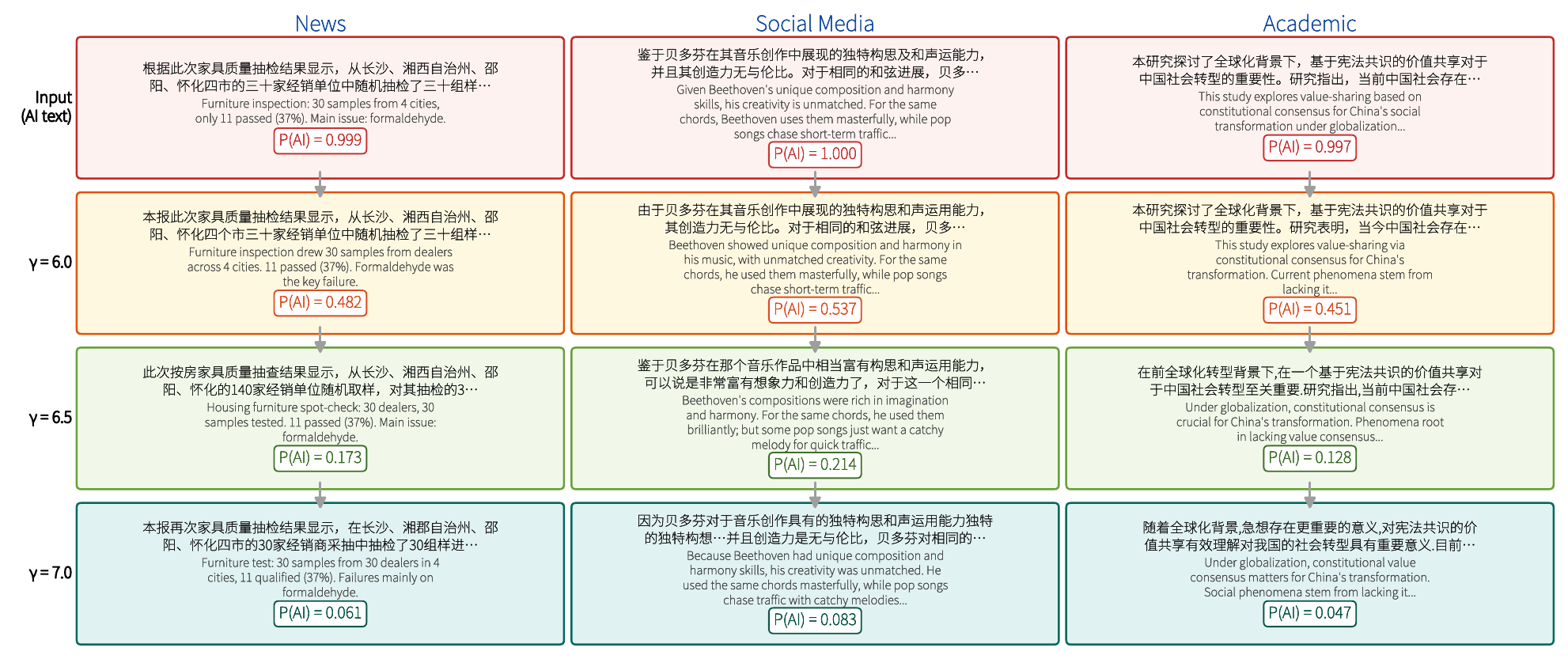}
    \caption{Qualitative examples across three domains at increasing $\gamma$.
    Each cell shows the Chinese text (top), an English gloss (middle), and the Det-v3 $\pai$ score (bottom).
    Row colors encode $\gamma$ intensity: \textcolor[HTML]{c62828}{red} (input), \textcolor[HTML]{e65100}{orange} ($\gamma{=}6.0$), \textcolor[HTML]{689f38}{light green} ($\gamma{=}6.5$), \textcolor[HTML]{00695c}{dark green} ($\gamma{=}7.0$).}
    \label{fig:qualitative}
\end{figure*}

Figure~\ref{fig:qualitative} presents one representative input--output example per domain (news, social media, and academic) at three transformation intensities, $\gamma\in\{6.0,6.5,7.0\}$.
Each cell shows the Chinese output, an English gloss, and the corresponding Det-v3 $\pai$ score; the top row is the original AI input.
Reading top to bottom, increasing $\gamma$ progressively rewrites formulaic AI phrasing into more colloquial language while preserving the core content, and the $\pai$ score drops accordingly---the per-example counterpart of the aggregate trends reported in \S\ref{sec:analysis}.

\section{Training Details}
\label{sec:app_training}

\noindent Data Construction Pipeline.\quad
For each domain, we collect human-written texts and use Qwen2.5-7B-Instruct with a domain-specific prompt to rewrite them into AI style.
We then score both sides of each pair with the Det-v3 AIGC detector and apply dual-end strict filtering: pairs are retained only if the human side scores $\pai < 0.1$ (confirming human-like style) and the AI side scores $\pai > 0.9$ (confirming strong AI characteristics).
This strict filtering ensures clean stylistic separation in every training pair.
The final dataset composition is:
\begin{itemize}
    \item Social Media: ${\sim}$346K pairs
    \item News: ${\sim}$68K pairs
    \item Academic: ${\sim}$22K pairs
\end{itemize}

\noindent Compute Resources.\quad
All training runs use 2 nodes of 8$\times$A800 (80\,GB) GPUs (16 GPUs total) with PyTorch DDP.
The full model (100K steps) takes approximately 48 hours; ablation variants (60K steps) take approximately 29 hours each.
Inference uses a single A800 (80\,GB) GPU; each 512-token sample takes ${\sim}$6 seconds (64 Euler steps).

\noindent Checkpoint Selection.\quad
We use a two-phase selection process: (1)~a quick sweep on 50 samples with $\gamma{=}6.5$ using the composite score $(1 - \pai) \times 0.6 + \text{Sim} \times 0.4$, and (2)~full evaluation of the best checkpoint across all $\gamma$ values and detectors.
For the main model, step 30{,}000 was selected.

\end{document}